\documentclass[letterpaper]{article} % DO NOT CHANGE THIS
\usepackage{aaai25}  % DO NOT CHANGE THIS
\usepackage{times}  % DO NOT CHANGE THIS
\usepackage{helvet}  % DO NOT CHANGE THIS
\usepackage{courier}  % DO NOT CHANGE THIS
\usepackage[hyphens]{url}  % DO NOT CHANGE THIS
\usepackage{graphicx} % DO NOT CHANGE THIS
\usepackage{natbib}  % DO NOT CHANGE THIS AND DO NOT ADD ANY OPTIONS TO IT
\usepackage{caption} % DO NOT CHANGE THIS AND DO NOT ADD ANY OPTIONS TO IT
\usepackage{algorithm}
\usepackage{algorithmic}
\usepackage{enumitem}

\usepackage{newfloat}
\usepackage{listings}
\DeclareCaptionStyle{ruled}{labelfont=normalfont,labelsep=colon,strut=off} % DO NOT CHANGE THIS
\floatstyle{ruled}
\newfloat{listing}{tb}{lst}{}
\floatname{listing}{Listing}
\usepackage{booktabs}
\usepackage{multirow}

\title{ECLAIR: Enhanced Clarification for Interactive Responses}
\author{
    John Murzaku\thanks{Work done while interning at Adobe.}$^1$,
    Zifan Liu$^2$,
    Md Mehrab Tanjim$^3$,
    \\
    Vaishnavi Muppala$^2$,
    Xiang Chen$^3$,
    Yunyao Li$^2$
}
\affiliations{
    $^1$Stony Brook University\\ $^2$Adobe\\ $^3$Adobe Research\\
    jmurzaku@cs.stonybrook.edu, \{zifanl, tanjim, mvaishna, xiangche, yunyaol\}@adobe.com
}

\begin{document}

\maketitle

\begin{abstract}
We present ECLAIR (\underline{\textbf{E}}nhanced \underline{\textbf{CLA}}rification for \underline{\textbf{I}}nteractive \underline{\textbf{R}}esponses), a novel unified and end-to-end framework for interactive disambiguation in enterprise AI assistants. ECLAIR generates clarification questions for ambiguous user queries and resolves ambiguity based on the user's response.We introduce a generalized architecture capable of integrating ambiguity information from multiple downstream agents, enhancing context-awareness in resolving ambiguities and allowing enterprise specific definition of agents. We further define agents within our system that provide domain-specific grounding information. We conduct experiments comparing ECLAIR to few-shot prompting techniques and demonstrate ECLAIR's superior performance in clarification question generation and ambiguity resolution. 
\end{abstract}

% Uncomment the following to link to your code, datasets, an extended version or similar.
%
% \begin{links}
%     \link{Code}{https://aaai.org/example/code}
%     \link{Datasets}{https://aaai.org/example/datasets}
%     \link{Extended version}{https://aaai.org/example/extended-version}
% \end{links}

\section{Introduction}
In language and conversation, humans naturally detect and resolve ambiguities by asking contextually relevant clarification questions and integrating information from multiple linguistic and paralinguistic cues. Recently, large language models (LLMs) have made significant strides in dialogue understanding tasks, demonstrating some ability to detect ambiguity and then ask clarification questions \cite{kuhn2022clam,zhang2023clarify,zhang2024clamber}. 

Despite these advancements, recent prompt-based LLM approaches and ambiguity corpora face several key limitations. These recent LLM approaches use a sequential pipeline: they first detect ambiguity, then generate a clarification question when needed. The previous approaches fail to capture the interplay and interdependence between ambiguity types and targeted clarification question generation. Furthermore, these systems treat ambiguity detection as a purely lexical problem, failing to incorporate contextual information from other linguistic cues or enterprise and domain-specific knowledge (i.e., enterprise knowledge graphs, classifiers, or databases). This narrow focus limits their ability to understand and resolve ambiguities in real-world, context-rich scenarios, particularly in enterprise settings where domain knowledge is critical and access to broader datasets and context is readily available. 

Regarding ambiguity and clarification question datasets, previous approaches oftentimes lack the complexity and nuance of real-world queries encountered in real industry applications. Like the systems, the datasets fail to capture the diverse range of ambiguities and contextual dependencies that arise in practical, domain-specific scenarios, which further limit their generalizability to real-world use cases. Furthermore, recent datasets are synthetic; while synthetic datasets  can be useful for dataset curation at scale, they cannot fully replicate the authenticity, unpredictability, and naturalness of real-world AI assistant interactions and queries from business users. 

Our work focuses specifically on the Adobe Experience Platform (AEP) AI Assistant \citep{bhambhri2024ai}. The AEP AI Assistant is a generative AI assistant built
for an enterprise data platform, where business users interact
with the system via a conversational interface to obtain answers based on multiple data sources. To motivate our system, we provide the following example: given a query ``ABC", while traditional LLM approaches to ambiguity detection and clarification questions might simply ask ``What do you mean by ABC?", our system approach, leveraging specific agents, refines this by asking, "Do you mean the dataset or the segment?", demonstrating a more nuanced understanding of our AEP AI Assistant enterprise-specific contexts. 

There are four main contributions of this work:

\begin{enumerate}[label=\roman*.]
   
\item We introduce a unified prompt-based approach that generates both ambiguity decisions and clarification questions in a single pass, improving efficiency over previous sequential methods.

\item We present a generalized architecture that can integrate ambiguity information and domain-specific knowledge from multiple downstream agents, enhancing context-awareness and flexibility in detecting and resolving ambiguities.

% \item We present our procedure for our internal dataset ECLAIR-CQ, a clarification question dataset annotated in-house on real user interactions in the Adobe Experience Platform (AEP) AI assistant. Our in-house annotation procedure explicitly captures the nuances of natural dialogue with AI assistants, specifically incorporating conversational context, interaction history, and grounding on specific AEP actions.

\item We evaluate ECLAIR on real user data and show that our unified approach performs better than standard prompting methods used in previous approaches.
\item We share our experiences and key takeaways in developing a disambiguation module for domain-specific enterprise AI assistants.
\end{enumerate}

\begin{figure*}
    \centering
    \includegraphics[width=0.8\linewidth]{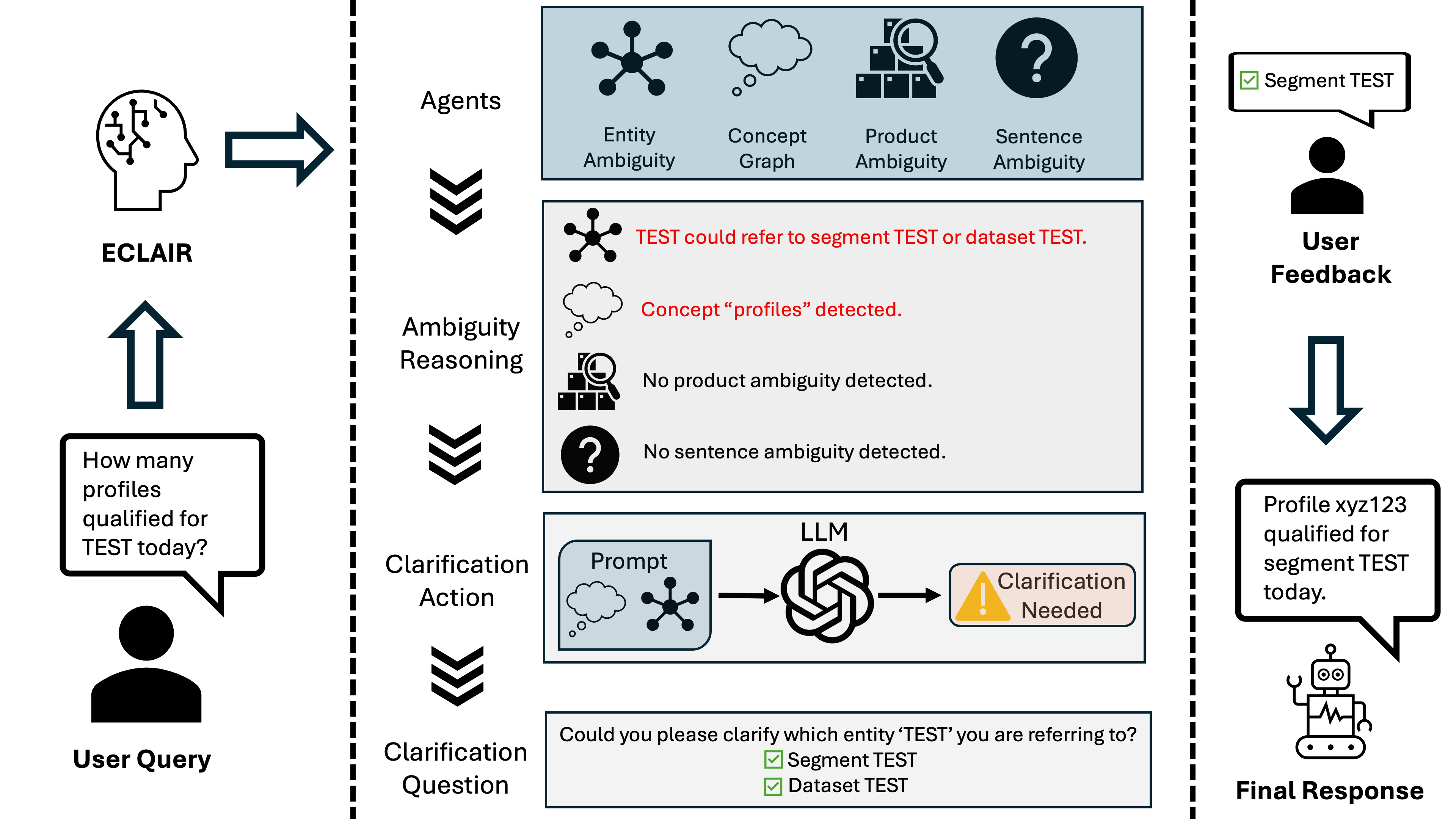}
    \caption{ECLAIR architecture diagram.}
    \label{fig:eclair-diagram}
\end{figure*}

In the following sections, we first briefly walk through the previous work on ambiguity detection and clarification question generation, particularly focusing on recent LLM prompt based approaches and their respective corpora. We then describe our implementation of ECLAIR and its key components, showcasing its modularizable and generalizable architecture. We describe our dataset collected and annotated on AEP AI assistant customer data, followed by baseline experiments on it. Finally, we share our lessons and experiences in developing ECLAIR, and the key steps necessary for deploying ECLAIR into production.
%%To change// placeholder for now/

\section{Related Work} 
% -> CQ datasets, AMBI classifiers, shortcomings, why we are better

\begin{table*}[]
\centering
\resizebox{\textwidth}{!}{
\begin{tabular}{p{3cm}p{6cm}p{7cm}}
\toprule
\textbf{Type} & \textbf{Example} & \textbf{Explanation} \\ 
\midrule
{Contextual} & \textit{``Can you show me the schema of this dataset''} &  It is not clear what the user is referring to by ``this''.\\ \hline
{Syntactic} & \textit{``XYZ123''} &  The user's intent about ``XYZ123'' is unclear.\\ \hline
{Aleatoric (Where/When/What)} & \textit{``Show me segments over time''} &  It is not clear what time span ``over time'' referring to.\\ 
\bottomrule
\end{tabular}}
\caption{Examples of ambiguous queries present in the AEP AI Assistant queries.}
\label{tab:examples}
\end{table*}
\subsection{Corpora}
Many corpora explore the notion of ambiguity and the need for clarification questions including: CLAQUA \citep{xu-etal-2019-asking}, ClariQ \citep{aliannejadi2020convai3}, Abg-CoQA \citep{guo2021abgcoqa}, Ambiguous TriviaQA \citep{kuhn2022clam}, PACIFIC \citep{deng2022pacific}, and CLAMBER \citep{zhang2024clamber}. These corpora differ along
several dimensions; we list the dimensions which are relevant for this paper.

(1) The specific ambiguity types captured. For example, Ambiguous TriviaQA focuses solely on entity ambiguities (pronoun or generic noun). In contrast, corpora like Abg-CoQA and PACIFIC not only include entity ambiguities (coreference), but also both include time ambiguities (i.e., the time frame a question refers to), specific answer type ambiguities (e.g., user wanted a general answer instead of a specific), and event ambiguities (e.g., multiple similar events mentioned). 

(2) The specific annotation strategies and taxonomies. For instance, CLAQUA's annotation procedure require a comprehensive multi-step process, including ambiguous sub-graph extraction from a knowledge base, ambiguous question annotation, and clarification question annotation. This comprehensive process aims to fully capture the process from ambiguity detection supplemented by a knowledge base, to a final clarification. ClariQ on the other hand employs a crowdsourcing annotation strategy, where each interaction is then labeled from a Likert 1-4 scale for necessity of clarification. These differences in annotation strategies influence the granularity, dataset sizes and scales, and types of ambiguities captured.

(3) The diversity of data sources. The corpora range from naturally occurring dialogues found in real-world data or applications, to templated structures and scenarios in a controlled environment. For examples, Ambiguous TriviaQA draws from the TriviaQA corpus \citep{joshi-etal-2017-triviaqa}, where many questions are naturally occurring entity related questions. As a result, the corpus contains only entity related questions in a trivia genre. In contrast, datasets like ClariQ and CLAQUA are built using templated approaches that systematically generate ambiguous scenarios designed to trigger the need for clarification. These templates ensure consistency in the types of ambiguities covered, such as entity disambiguation, request specificity, and temporal aspects, allowing for a structured and predictable dataset. 

%%Table from Mehrab's paper/

\subsection{Methods}
Recent papers on ambiguity detection and clarification question generation use prompt-based LLM approaches \citep{kuhn2022clam,deng-etal-2023-prompting,zhang2024clamber}. These approaches follow standard zero-shot or few-shot chain-of-thought (CoT) prompting methods. These methods are typically used as a pipeline: a LLM is first prompted to determine if a question is classified as ambiguous or not ambiguous, and if it is, the LLM is again prompted to generate a clarification question. Previous literature has found that LLMs are typically inadequate at resolving ambiguities (with relatively low performance metrics) and fail to generate meaningful clarification questions due to knowledge gaps. To explore this issue, CLAMBER was proposed as the first generalized benchmark, further showcasing the inadequacies and difficulties of LLMs in generalized ambiguity detection and clarification question generation. Outside of prompt based approaches, \citet{zhang2023clarify} propose an uncertainty estimation technique for ambiguity detection, aiming to capture the entropy over user intents by simulating multiple user-assistant interactions.

Our work differs from the related work by offering two salient contributions: first, we test on a corpus based on real-world interactions with AEP AI Assistant, which includes a comprehensive coverage of ambiguity types present in natural conversations. 
Our second contribution is in our methodology: we present a unified prompt-based approach, which integrates ambiguity information from multiple downstream agents for a comprehensive overview of ambiguity.

\section{ECLAIR Overview} 
% -> architecture diagram; 1,2,3 of current components 
The ECLAIR architecture (Figure 1) is built as a modularizable and extendable framework: the AI Assistant developers are able to fully define their custom ambiguity detection agents or grounding information agents. Our system allows multiple ambiguity agents. We describe a step-by-step sample interaction and what happens in our system.

Each interaction begins with a user query. The system then initializes a set of predefined ambiguity detection agents, which are predefined and configured by the user. These agents are modular and can be tailored to specific types of ambiguities, making the framework extensible and customizable. 
In our specific use case, we define the following agents:
\begin{itemize}
    \item A generic sentence-level ambiguity detector that captures domain-agnostic ambiguities.
    \item A product ambiguity detector that decides if the query refers to multiple products in Adobe AEP.
    \item A entity linking ambiguity detector that identifies spans of text that can be linked to multiple entities in the database.
    \item A concept graph module that identified tokens that are AEP-specific terminology.
\end{itemize}
% a generic sentence-level ambiguity detector, then specialized ambiguity detectors such as a product classifier to classify if there is a product ambiguity; entity linking to detect if an entity refers to multiple entities; a concept graph to ground our agent into Adobe Experience Platform specific terminology; 
% a classifier called Need2Rewrite which detects lexical ambiguity; 
% and finally, a Question Rewrite agent, which rewrites a query if ambiguity exists. 

Next, each agent processes the query to determine if ambiguity or domain-specific terminology is present. For example, an entity linking agent would analyze the query to assess whether a mentioned entity is clearly defined or if it maps to multiple possible entities in the database, indicating ambiguity. Similarly, if a query can be answered in the context of multiple products, this agent would flag an ambiguity as present. The concept graph module is an Adobe AEP grounding agent which does not return ambiguity; rather, it grounds our prompt onto our domain. We will further describe this grounding process later.

If an agent identifies an ambiguity (e.g., the entity linking agent returns multiple potential matches, or the generic ambiguity detector detects lexical ambiguity), the system incorporates the agents' output into a prompt. For each agent that gives an output, the following will be added to the prompt: a description of the agent, and the detail of the ambiguity, where both can be customized per agent. This step ensures that all detected ambiguities from the downstream agents are represented in the prompt that will be used in the next stage. In the specific case of our Entity Linking agent, the ambiguity in the prompt will be added as: 
\begin{itemize}
    \item \textbf{Agent description:} Entity linking in serves the function of associating or connecting mentions of entities to the corresponding entries in the database.
    \item \textbf{Ambiguity Detected:} \texttt{True}
    \item \textbf{Matches:} \texttt{ENT} can be linked to \texttt{ENT\_1}, \texttt{ENT\_2}, \texttt{ENT\_3} 
\end{itemize}

The constructed prompt containing the outputs from all the defined agents is then passed into a LLM, with instructions on detecting ambiguities and generating clarifications. The LLM then evaluates the information from the overall prompt and first makes a binary decision: whether the query contains ambiguity or not. If ambiguity is detected, the LLM generates a clarification question aimed at resolving the detected ambiguity. This question is designed to elicit further input from the user to clarify their intent, referencing the specific agents as well. We also include grounding information in the prompt to help the LLM understand the domain specific context: this can include keywords, enterprise specific instructions, and definitions of enterprise terms.

Our prompt approach has two main benefits. First, it is modularizable: users can ground to their specific platform and define custom agents. Second, it unifies the ambiguity detection and clarification question generation in a singular LLM call, effectively capturing the interdependence between the type of ambiguity and the formulation of the clarification question. 
% We note that we performed experiments with a pipeline decision (i.e., first prompt the LLM for ambiguity detection enhanced with the agents output, then prompt for a clarification question), but note that this performed worse.
% Note that We conducted experiments using a pipeline approach, where we first prompted the LLM for ambiguity detection (enhanced with the agent's output) and then prompted for a clarification question. However, this approach yielded poorer results.

\begin{table*}[]
\centering
\resizebox{\textwidth}{!}{
\begin{tabular}{p{10cm}p{6cm}}
\toprule
\textbf{Query} & \textbf{Annotation} \\ 
\midrule
\textit{``Confirm if data is currently being ingested''} & Could you please specify which data you are referring to and where it is being ingested?\\ \hline
\textit{``abc123''} & Could you kindly specify what you would like to know about ``abc123''?\\ \hline
\textit{``Which sandbox is this referencing?''} & Could you kindly clarify what you mean by ``this''?\\ \hline
\textit{``Which segment is not in use?''} & None needed.\\ \hline
\textit{``What is my largest audience''} & None needed.\\ \hline
\textit{``list of fields in abc123''} & None needed.\\ 
\bottomrule
\end{tabular}}
\caption{Examples of user queries and their corresponding annotations.}
\label{tab:query_annotation}
\end{table*}

\section{Experiments}
We evaluate ECLAIR on a sample of real user queries and demonstrate that ECLAIR improves the precision in determining whether to ask a clarification question by 17 points compared to a standard few-shot prompting method. The higher precision helps avoid unnecessary clarification questions, which can otherwise negatively impact user experience. Furthermore, the clarification questions generated by ECLAIR aligns better with gold questions generated by human than the baseline.

\subsection{Experimental Setup}
We introduce the few-shot prompting baseline method that is built upon the current AEP AI Assistant architecture. Then we introduce the evaluation data and the annotations.

\paragraph{Baseline} 
We develop a few-shot prompting method based on the current AEP AI Assistant pipeline as the baseline method.
In the current pipeline, there are two modules for disambiguation: a ambiguity detector which is a binary classifier and a query rewriting module that rewrites the query based on the chat history to resolve contextual ambiguities.
We build our few-shot baseline by attaching the same ambiguity detector after query rewriting. If the query is still classified as ambiguous, we prompt GPT 3.5 to generate clarification questions is needed. 10 human-crafted examples are provided in the prompt as the few shot examples.

\paragraph{Evaluation Dataset} 
Our evaluation dataset consists of a sample from real user queries in the AEP AI Assistant.
We filter out examples that are classified as unambiguous using the ambiguous detector after query rewriting. 
We are left with 100 queries that are hard cases where the ambiguity cannot be resolved using chat-history-based query rewriting.
Among the 100 queries, most of the ambiguity types fall under three categories: contextual, where the context or reference to an object is underspecified; syntactic, where the sentence is malformed or incomplete, leading to indirect interpretation; and aleatoric, where a specific token is undefined or contains multiple possible meanings.Table~\ref{tab:examples} shows the ambiguity types, the anonymized real user queries, and an explanation of why it is ambiguous. 

\paragraph{Annotation} Annotators are given the following: the ambiguous query, the rewritten verions of the query from the query rewriting module and the chat history between the AI assistant and the user. Annotators are asked to first determine if a question needs clarification or not. If clarification is not needed, the annotators annotate ``None needed''. If clarification is needed, the annotators write the clarification question as succinctly as possible.  We randomly split the annotation assignment among four annotators, ensuring that each example had exactly three annotations. If disagreement on the need for a clarification question was present (i.e., two annotators said ``None needed'', while one wrote a clarification question, or vice versa), we assigned that example to one of the four annotators who did not see it. We then performed a majority vote, choosing whichever label was the majority. If further disagreement was present again (which in this case, would be two annotators annotating ``None needed'' and two annotating a clarification question), we assigned the example to a fifth annotator to resolve the disagreement.

\paragraph{Annotation Agreement} We analyze the agreement between the annotators' clarification questions. Our primary focus is semantic meaning and structure: we evaluate the annotator's clarification questions on BERTScore \citep{zhang2019bertscore} for semantic similarity, and ROUGE-L for structural similarity. Among our annotators, we achieve an average BERTScore of 0.904 and an average ROUGE-L of 0.395. The high average BERTScore indicates strong semantic similarity between annotators' clarification questions, while the average ROUGE-L suggests moderate overlap in sequence, highlighting alignment in structural consistency.

\paragraph{Examples of Annotations} Table~\ref{tab:query_annotation} shows three examples of queries needing clarification and three examples of no clarification needed. We note the annotations were either very similar or identical among annotators in these cases (if not identical, minor differences such as quotation marks around entities or punctuation).

% \paragraph{Discussion} While our dataset is small, we summarize our salient annotation and dataset contributions in the ambiguity detection and clarification question generation tasks:
% \begin{itemize}
%     \item \textbf{Real user data} ECLAIR-CQ is built on top of real user queries in a production system.
%     \item \textbf{Comprehensive annotation procedure} Our annotators follow a two-step annotation procedure: annotate whether a clarification question is needed or not, and then if yes, ask the clarification question. We further refine our annotations until agreement is reached by recruiting further annotators. 
%     \item \textbf{Token-level annotations} Our annotators further annotate ambiguous tokens. As mentioned, this will be used in future work to pinpoint exactly where ambiguity occurs.
% \end{itemize}

% -> dataset, annotation, corpus sample
\begin{table*}[h]
    \centering
    \begin{tabular}{lcccccc}
        \toprule
        & \multicolumn{3}{c}{\textbf{Baseline}} & \multicolumn{3}{c}{\textbf{ECLAIR}} \\
        \cmidrule(lr){2-4} \cmidrule(lr){5-7}
        & Precision & Recall & F1 & Precision & Recall & F1 \\
        \midrule
        Clarification Needed & 0.732 & \textbf{0.833} & \textbf{0.779} & \textbf{0.904} & 0.635 & 0.746 \\
        Clarification Not Needed & 0.333 & 0.214 & 0.261 & \textbf{0.438} & \textbf{0.808} & \textbf{0.568} \\
        Average & 0.533 & 0.524 & 0.520 & \textbf{0.671} & \textbf{0.721} & \textbf{0.657} \\
        \bottomrule
    \end{tabular}
    \caption{Comparison of Precision, Recall, and F1 Score for Old method (Few-shot) and ECLAIR. Improvements are shown in bold.}
    \label{tab:comparison}
\end{table*}

\paragraph{Evaluation Protocol} We perform two evaluations: first, we perform an evaluation on whether a clarification is needed or not. We evaluate this with standard Precision/Recall/F1 metrics. We note that for our production system, our major goal is to maximize precision; in other words, we only want to ask clarification questions when it is absolutely necessary, and not affect the overall user experience. We also perform a BERTScore evaluation to evaluate how aligned our system's clarification generations are with the gold annotated examples. 
\subsection{Experimental Results}  

\paragraph{Binary Decision} Table 3 shows the results for ECLAIR compared to our few-shot baseline on deciding whether clarification question is needed or not. We first observe that ECLAIR achieves higher precision on ``Clarification Needed'', but falls short on recall and F1. However, we emphasize that for our production setting, we are aiming to maximize precision (that is, we do not want to over-clarify, and only ask clarifications when they are needed). ECLAIR yields major improvements on the CQ Not Needed label: we see a boost in precision, recall, and F1, with the largest boost occurring in recall. In turn, ECLAIR yields a overall F1 that is 13\% higher than our few-shot method. This emphasizes that our agentic approach not only captures more information and provides a clearer picture of ambiguity, but also yields better metrics compared to standard few-shot methods. 

\paragraph{Clarification Question Generation} We conduct a BERTScore evaluation to assess how closely our system's clarification generation aligns with the gold-annotated examples. The BERTScore for the few-shot baseline is 0.868. For our ECLAIR approach, the BERTscore was higher, yielding an average BERTScore of 0.886. Our results show two key findings: first, our gold annotated dataset and the output of GPT-3.5 are semantically similar, showing the validity of using a prompt-based LLM approach. Second, we show that ECLAIR yields a slightly better BERTScore than the baseline few-shot method, aligning better with our gold annotations.
%-> eval, few-shot vs ECLAIR, F1, <user_study>
\section{ECLAIR: Two Production Use Cases}
In this section, we showcase two use cases of the ECLAIR system on our AEP AI Assistant UI. We show ECLAIR helping disambiguate a query with an ambiguous entity and ECLAIR helping disambiguate a query that can refer to multiple Adobe products. We show further additions to our base AI assistant UI, where we added clickable buttons that help aid the user in the disambiguation process. Once the user clicks the button, the system uses that button's label as the clarification and returns a correct answer.
\begin{figure}
    \centering
    \includegraphics[width=\linewidth]{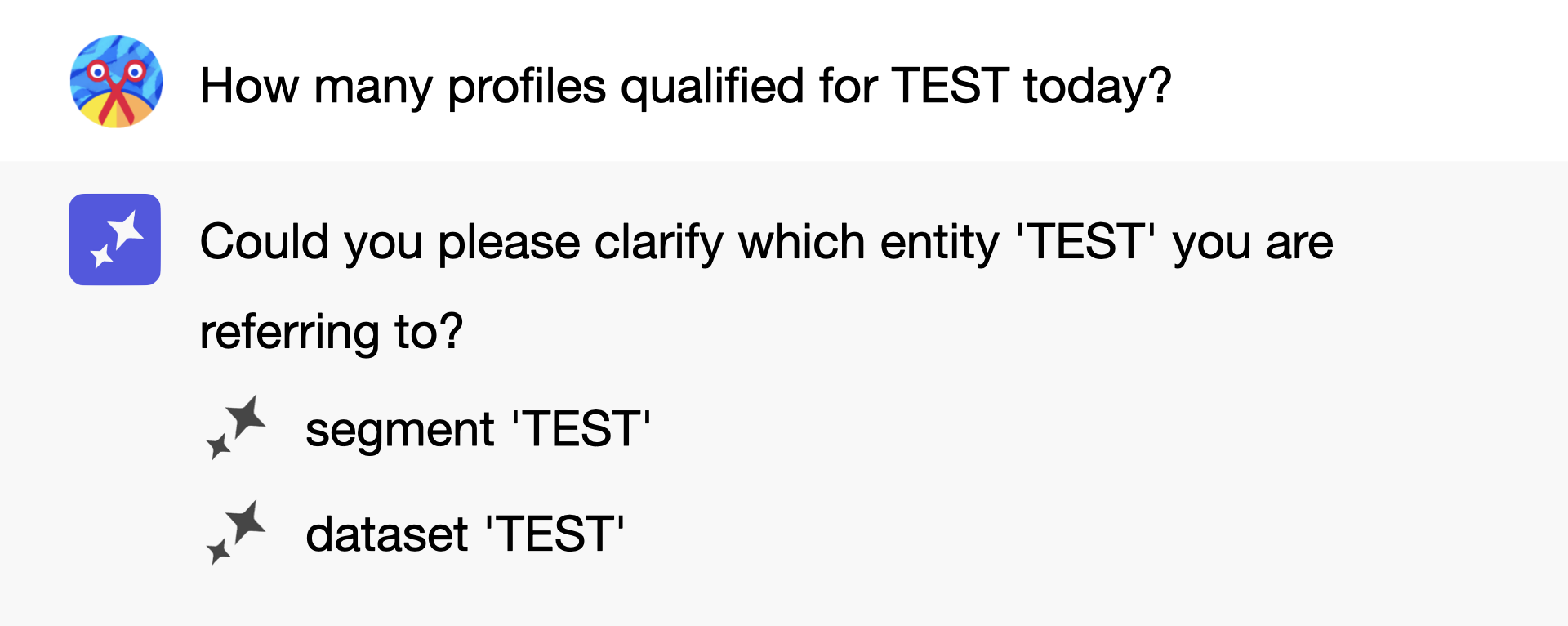}
    \caption{Entity ambiguity shown in our UI. The options are clickable buttons the user can interact with.}
    \label{fig:ent-amb}
\end{figure}

\paragraph{Use Case 1: Ambiguous Entity} Figure 2 shows a case where a user refers to an entity called TEST. Here, ECLAIR successfully shows that there is an entity ambiguity (through the entity linking agent), and generates a useful clarification question. The user then can clarify which specific entity type they are referring to: a segment called TEST or a dataset called TEST.

\paragraph{Use Case 2: Ambiguous Product} Figure 3 shows a case where a users product query can refer to multiple Adobe products. ECLAIR successfully captures that this query is ambiguous (from the product classifier agent), and asks a relevant clarification question. Similar to our first use case, the UI displays three button choices for the user: Adobe Workfront, Adobe Experience Manager, and Adobe Commerce. 
\begin{figure}
    \centering
    \includegraphics[width=\linewidth]{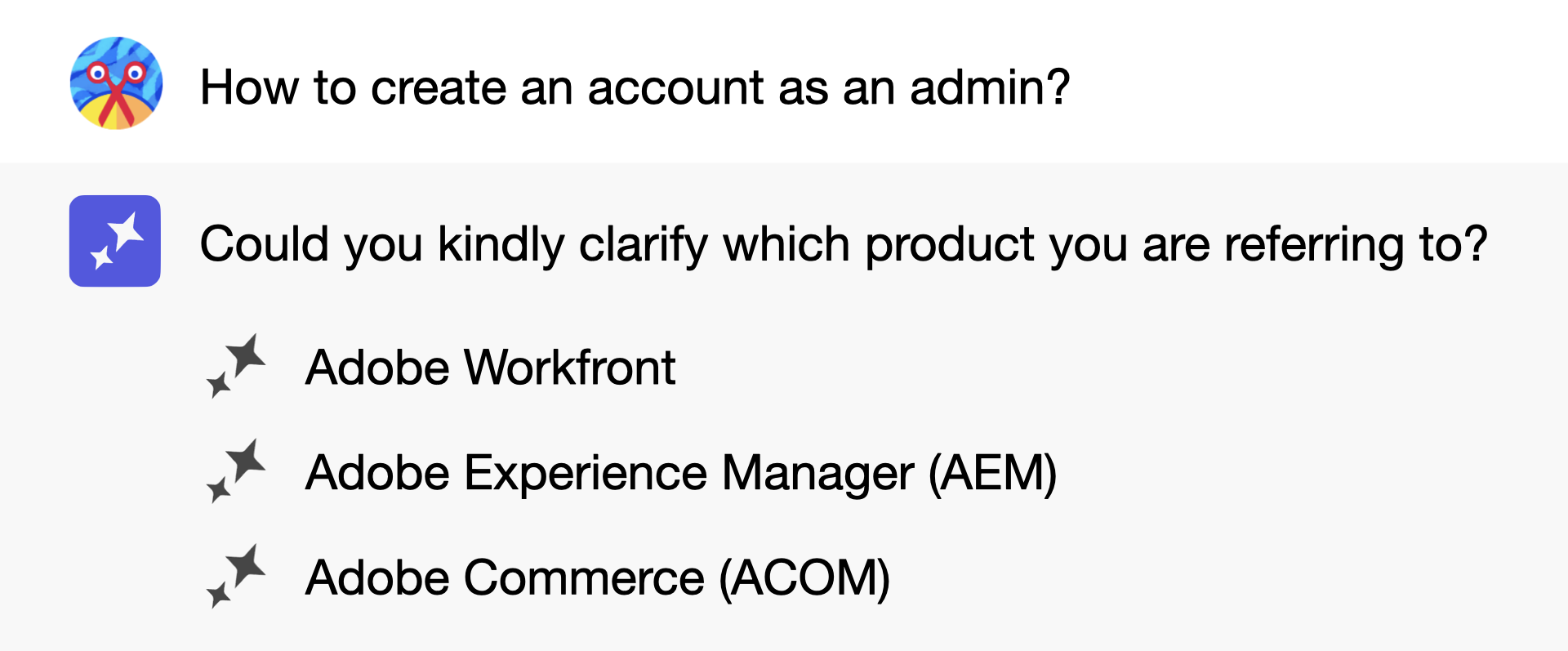}
    \caption{Product ambiguity shown in our UI. The options are clickable buttons the user can interact with.}
    \label{fig:prod-amb}
\end{figure}

We also performed experiments on when to surface clarification questions for the best user experience. For example, in some cases we first surface the answer to a product question, but also inform the user of a potential ambiguity. Figure 4 shows this where the answer is first shown, but at the end of the query, we surface the ambiguity to the user so that they could follow up if needed. We will explore this in future work and add robust clarification strategy decisions in future iterations of ECLAIR.

\section{Experiences and Lessons}
In this section, we share the key lessons learned and practical insights gained from our experience in developing ECLAIR:
\begin{itemize}
    \item For domain-specific AI assistants, it is essential to provide related domain knowledge if we use a general-knowledge LLM to ask clarification questions.
    \item From user interactions, we observe that users sometimes treat AI assistants like a search engine, entering only keywords. In these cases, inferring the user's intention often provides better user experience than asking clarification questions.
    \item The ambiguity of a question depends on the user's profile, as previous interactions and preferences may offer implicit information for disambiguation. In the future, we plan to design an agent that can retrieve relevant user profile information for more accurate disambiguation.
\end{itemize}

\section{Path to Deployment} 
%-> how we will integrate it, current pipeline; -> AJO integration
Deploying ECLAIR into production requires careful planning, extensive testing and significant efforts across multiple dimensions. We highlight the following key aspects that are essential for ensuring a successful and robust deployment and discuss our plans.
\begin{itemize}
    \item \textbf{Precision} It is crucial that clarification questions are asked only when necessary and they precisely target the ambiguity in the user's query, as unnecessary or inaccurate clarification questions can significantly disrupt the interaction and lead to negative user experience. Our goal is to achieve a precision rate of 98\% or higher, ensuring that whenever a clarification question is asked, it is both necessary and accurate in most of the cases. To avoid unnecessary or inaccurate clarification questions, we plan to integrate more domain-specific knowledge to ground the user query into the domain-specific knowledge. In addition, we will provide additional examples to illustrate clear versus unclear questions, and refine the LLM prompt to make it more conservative.
    \item \textbf{Latency} Since executing each agent and making LLM calls can be time-consuming, the time it takes for a user to receive a final response can significantly increase after the integration of ECLAIR. To manage the latency, we need to optimize the pipeline. We plan to identify dependencies between downstream agents and enable parallel execution whenever possible.
    \item \textbf{User Study} A broad user study is essential to evaluate the performance of ECLAIR in improving conversational experience and check if our design is aligned with user expectations. Once ECLAIR meets the precision and latency criteria, we plan to activate this feature for a representative sample of users and gather their feedback.
    \item \textbf{Continuous Monitoring and Iteration} Following the deployment of ECLAIR, continuous performance monitoring and iterative improvements are essential. We intend to track both explicit and implicit user feedback, such as the ratio of thumbs-up to thumbs-down, the percentage of clarification questions that result in users abandoning the conversation, and explicit user comments. Using the feedback, we will iteratively refine and enhance ECLAIR.
\end{itemize}

\begin{figure}
    \centering
    \includegraphics[width=\linewidth]{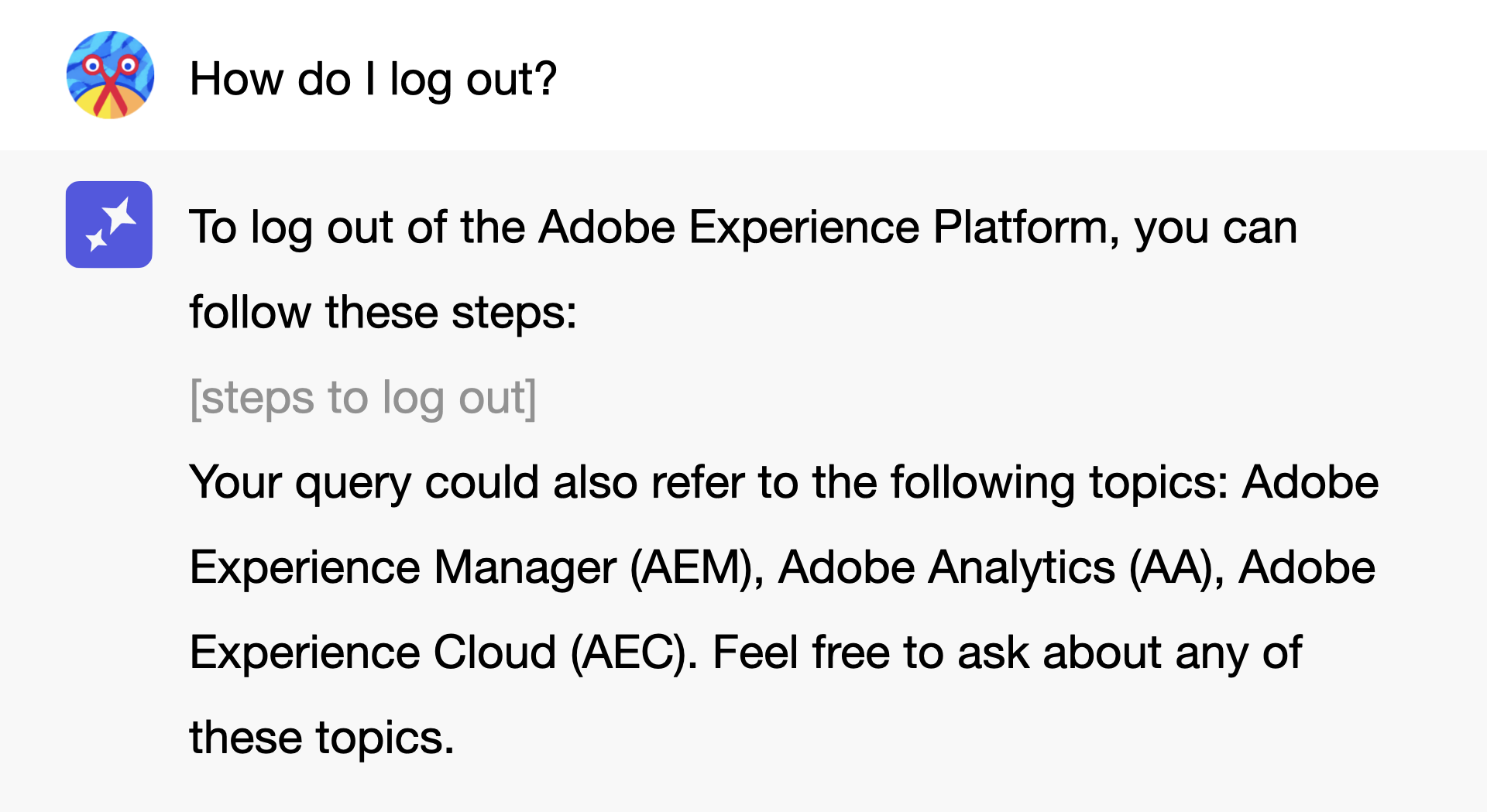}
    \caption{Our UI where we first answer the query, but also inform the user of potential ambiguity.}
    \label{fig:enter-label}
\end{figure}
\section{Conclusion} 
After introducing the ECLAIR framework, we demonstrate its superior performance in ambiguity detection and clarification question generation across a real-world dataset from the Adobe Experience Platform (AEP) AI Assistant. ECLAIR's modular and unified approach not only outperforms traditional few-shot methods (Table 3) but also integrates seamlessly into enterprise AI systems, addressing practical challenges such as latency and precision. Through the agentic architecture, ECLAIR enhances context-awareness and flexibility, showing significant improvements in both ambiguous entity resolution (Figure 2) and product disambiguation (Figure 3). Moving forward, we suggest refining ECLAIR’s grounding mechanisms and optimizing the pipeline to further reduce latency. Additionally, we note that conducting broader user studies will ensure that ECLAIR continues to align with user expectations. This in turn will pave the way for robust deployment in our AEP AI Assistant use case, but will also serve as a blueprint for continuous disambiguation enhancements in enterprise AI assistants.
%-> conclude; future work; 

\section{Acknowledgments}
We would like to thank Victor Soares Bursztyn, Cole Connelly, Nathenael Dereb, Rachel Hanessian, Jordyn Harrison, Sai Sree Harsha, Sai Jayakumar, Shun Jiang, Akash Maharaj, Danny Miller, Kun Qian, Pawan Sevak, Jordan Walker for their insightful discussions and valuable help throughout the development of this work. Their guidance and feedback greatly contributed to the quality of this research.

\bibliography{aaai25}

\end{document}